\documentclass[letterpaper, 10 pt, conference]{ieeeconf} 

\IEEEoverridecommandlockouts                              % This command is only needed if 
\usepackage[binary-units=true]{siunitx}
\usepackage{times}
\usepackage{amsmath,amssymb}
\usepackage{accents}
\usepackage{graphicx}
\usepackage{bm}
\let\labelindent\relax
\usepackage[inline]{enumitem}
\usepackage{array}
\usepackage{multicol}
\usepackage{multirow}
\usepackage{tabulary}
\usepackage{booktabs}
\usepackage{subfig}
\usepackage{todonotes}
\usepackage{cite}
\usepackage[hidelinks]{hyperref}
\usepackage{algorithm}% http://ctan.org/pkg/algorithm
\usepackage{algpseudocode}% http://ctan.org/pkg/algorithmicx
\usepackage{breqn}
\usepackage{tcolorbox}
\usepackage{mathtools}
\usepackage{cuted}
\usepackage{stfloats}
\usepackage{balance}

\renewcommand{\thetable}{\arabic{table}}

\allowdisplaybreaks

\DeclareMathOperator*{\mini}{min.}

\definecolor{light-gray}{gray}{0.95}
\definecolor{dark-gray}{gray}{0.5}
\definecolor{mygray}{gray}{0.75}
\newcommand{\BIN}{\begin{bmatrix}}
\newcommand{\BOUT}{\end{bmatrix}}

\pgfdeclarelayer{bg1}   
\pgfdeclarelayer{bg}  
\pgfsetlayers{bg1,bg,main}  

\definecolor{orange}{rgb}{0.99,0.69,0.07}
\definecolor{lightgray}{gray}{0.85}
\definecolor{light-gray}{gray}{0.95}
\definecolor{dark-gray}{gray}{0.5}

\tikzset{cross/.style={cross out, draw=black, minimum size=2*(#1-\pgflinewidth), inner sep=0pt, outer sep=0pt},
cross/.default={1pt}}

 \makeatletter
 \newcommand\fs@spaceruled{\def\@fs@cfont{\bfseries}\let\@fs@capt\floatc@ruled
   \def\@fs@pre{\vspace{5pt}\hrule height.8pt depth0pt \kern2pt}%
   \def\@fs@post{\kern2pt\hrule\relax}%
   \def\@fs@mid{\kern2pt\hrule\kern2pt}%
   \let\@fs@iftopcapt\iftrue}
 \makeatother

\makeatletter
\newcommand{\thickhline}{%
	\noalign {\ifnum 0=`}\fi \hrule height 1pt
	\futurelet \reserved@a \@xhline
}
\newcolumntype{"}{@{\hskip\tabcolsep\vrule width 1pt\hskip\tabcolsep}}
\makeatother

\title{\LARGE \bf Path Planning Under Uncertainty to Localize mmWave Sources}
\author{Kai Pfeiffer$^{1}$, Yuze Jia$^{2}$, Mingsheng Yin$^{2}$, Akshaj Kumar Veldanda$^{2}$, Yaqi Hu$^{2}$, Amee Trivedi$^{3}$, Jeff Zhang$^{4}$,\\
	Siddharth Garg$^{2}$, Elza Erkip$^{2}$, Sundeep Rangan$^{2}$, Ludovic Righetti$^{2}$%
	\thanks{$^{1}$Schaffler Hub for Advanced Research at NTU and School of Mechanical and Aerospace Engineering, Nanyang Technological University, Singapore}%
	\thanks{$^{2}$Tandon School of Engineering, New York University, New York, USA}%
	\thanks{$^{3}$University of British Columbia, Vancouver, BC, Canada}
	\thanks{$^{4}$Harvard University, Cambridge, MA, USA}
	\thanks{Part of this work was supported by New York University, NSF grants
		1936332, 1824434, 1833666, 1564142, 1925079, 1825993; NYU WIRELESS and its industrial affiliates; NIST grant 70NANB17H166; SRC; and
		a research grant from OPPO. This work is partly supported by the Schaeffler Hub for Advanced Research at NTU, under the ASTAR IAF-ICP Programme ICP1900093.}
}

\begin{document}
	
	\maketitle
	\thispagestyle{empty}
	\pagestyle{empty}
	
	\begin{abstract}%
		%The millimeter wave bands, which 5G can be classified as, are favourable in their high angular and temporal resolution which enables high precision localization for example in robot control. 
		In this paper, we study a navigation problem where a mobile robot needs to locate a mmWave wireless signal. 
		Using the directionality properties of the signal, we propose an estimation and path planning algorithm that can efficiently navigate in cluttered indoor environments.
		% Robot navigation in cluttered environments based on wireless signals is challenging due to the possibly high degree of signal reflection.
		% With this paper we provide tools to navigate robots to transmitters in cluttered environments using the noisy angle of arrival of the received signal. 
		We formulate Extended Kalman filters for emitter location estimation in cases where the signal is received in line-of-sight or after reflections.
		%
		% and first and $n$-th order non-line-of-sight case with the signal emitting transponder. 
		We then propose to plan motion trajectories based on belief-space dynamics in order to minimize the uncertainty of the position estimates. The associated non-linear optimization problem is solved by a state-of-the-art constrained iLQR solver. In particular, we propose a method that can handle 
		a large number of obstacles ($\sim 300$) with reasonable computation times.
		%  Obstacle avoidance is formulated as constraints which discards the necessity for expensive second order derivatives. Additionally, due to the usage of analytical partial derivatives for both the Bayesian filtering dynamics and the constraints, the path planner can handle a large number of obstacles ($\sim 300$) while computation times remain tractable. 
		We validate the approach in an extensive set of simulations. We show that our estimators can help increase navigation success rate and that planning to reduce estimation uncertainty can improve the overall task completion speed.
		
		% and  and are successful in navigating the robot to previously unknown transponder positions.
	\end{abstract}
	
	%\begin{keywords}%
	%  List of keywords%
	%\end{keywords}
	
	\section{Introduction}
	Wireless communications play an important role in robotics, typically to operate robots~\cite{Tolley2014} or drones~\cite{Li2021} remotely.
	Additionally, wireless signals can be used as sensors, for example to localize a robot 
	or to perform SLAM based on the signal strength of a WiFi network~\cite{ferris2007}.
	However, traditionally wireless signals do not contain direction information to estimate the location of the emitter precisely. 
	%
	% in robotics this wireless connection oftentimes serves as a mean to an end rather than being incorporated as valuable information itself, for example for localization. Especially in search and rescue missions the robot is controlled wirelessly~\cite{bhondve2014}. The augmentation of measurements like vision or LiDAR with wireless data, for example the angle of arrival (AoA) or signal strength, can lead to higher SLAM accuracy.
	%
	%We want to make a step in this direction and consider the scenario of localizing a transmitter of a wireless signal in a cluttered environment based on the AoA. 
	The emergence of directional millimeter wave (mmWave) bands, which are deployed for example in 5G networks, have attracted significant attention for high precision localization
	applications. Indeed, the estimation of the direction of transmission is greatly simplified as angular information does not need to be reconstructed by complex algorithm as in~\cite{Zhang2008,Graefenstein2010,Zwirello2012}.
	
	In this paper, we study a scenario where a robot seeks to navigate to fixed but unknown mmWave wireless emitter location in an unknown environment.
	%and find the unknown location of a fixed mmWave wireless emitter. 
	%This problem can represent, 
	This could model, for example, a search and rescue scenario where the stranded human carries the emitter~\cite{Silvagni2017,Huafeng2018,Atif2021,Barry2021}. Our goal is to exploit the directionality of mmWave wireless to devise a planning and estimation framework for fast navigation to the emitter.
	
	%This work builds on our 
	Recent work~\cite{yin2022millimeter} has addressed this problem using a simple 
	%by classifying
	%studied how mmWave wireless signals could be processed and 
	%classified to estimate if they came directly from an emitter
	%in line of sight (LOS) or if the signal had been reflected before (NLOS). The work proposed a simple 
	navigation algorithm that follows the angle of arrival (AoA) of the received
	mmWave signal till the robot reaches the emitter. The algorithm can be used if the robot is in line of sight (LOS) or in non-line of sight (NLOS) settings of the emitter. 
	However, this method has two drawbacks. First, it uses noisy, unfiltered AoA measurements and is thus ineffective when the robot is two or more reflections from the emitter. Second, it does not maintain an estimate of the transponder position, requiring the robot to navigate first to the point of reflection in NLOS settings.
	%preventing direct navigation to the emitter location because the robot instead navigates to the point of reflection in NLOS.
	%which might be required, for example, if signal loss needs to be avoided.
	
	% In this paper, we proposed to extend the system with an estimator of the position
	% of the emitter and planning algorithm
	% that aims to reduce the uncertainty of the emitter position while navigating towards it.
	
	%In this vein, the authors in~\cite{Graefenstein2010} proposed a localization algorithm based on a particle filter and an Extended Kalman filter (EKF) in an outdoor scenario with LOS with the emitter. 
	
	This paper aims to improve upon the state-of-art using joint estimation and trajectory planning algorithms. We consider unexplored cluttered environments with possibly high degrees of signal reflections (NLOS). We propose an Extended Kalman Filter (EKF) to estimate the robot position with respect to the emitter in the LOS and NLOS cases. The filters are able to robustly estimate the direction of the wireless sources despite noisy observations. 
	We formulate a trajectory optimization problem based on belief-space dynamics following~\cite{berg2011} to minimize the uncertainty of the estimate along the trajectory while avoiding obstacles.
	
	We propose an improved algorithm compared to \cite{berg2011} by formulating the trajectory optimization problem as a constrained one. This renders the need for expensive second-order derivatives of the cost terms when including collision avoidance unnecessary. Solvers have been proposed that are able to consider box constraints~\cite{Tassa2014}, or nonlinear inequality constraints for example by the augmented Lagrangian method~\cite{Jackson2021} or the interior-point-method~\cite{pavlov2021}. Furthermore, we provide explicit computations of the partial derivatives of the estimation covariance updates of the EKF, necessary for the solvers. This brings computational advantage compared to numerical derivatives previously reported~\cite{berg2011}. 
	The proposed method is evaluated using the 'Active Neural-SLAM' framework\cite{chaplot2020}, a modular, open-source, approach for mobile navigation, 
	%We integrate our approach in the Neural-SLAM navigation framework 
	and demonstrate its performance benefits over prior work in a wide range of simulated indoor navigation environments.
	
	% The authors in~\cite{chaplot2020} proposed a learning based SLAM modality referred to as `Active Neural-SLAM'. With its modular structure the algorithm is able to efficiently explore unknown real-world environments while being flexible with regards to the sensor inputs and robust with respect to state estimation errors. In our algorithm we rely on this module to obtain a global robot position estimate in a 2D obstacle map and for exploration in areas of no reception.
	
	In the following, we formulate the EKF transition and observation models for the LOS and first- and general n-th order NLOS cases (Sec.~\ref{sec:EKF}). To obtain trajectories that minimize the uncertainty of the EKF state estimate we formulate a non-linear constrained trajectory optimization problem (Sec.~\ref{sec:path}). Based on Bayesian filtering, we require the gradient of the EKF which is detailed in App.~\ref{app:ekfgrad}. We start the evaluation (Sec.~\ref{sec:eval}) by verifying the identification based on the EKF (Sec.~\ref{sec:evalID}) and path planning through cluttered environments (Sec.~\ref{sec:evalPP}) in simple scenarios. The methods are then applied on complex maps,  with detailed results on a single typical map (Sec.~\ref{sec:evalPP}), and success rate and time-to-target evaluations on 200 different scenarios (Sec.~\ref{sec:benchmark}).

	\section{Extended Kalman filter for transponder localization}
	\label{sec:EKF}
	
	In this section we describe the filter used to estimate the location of the 
	robot with respect to the wireless emitter. We need to distinguish several cases depending
	on whether the received signal comes from an emitter in LOS, in first-order NLOS (i.e. 
	when there was only one reflection) or in n-th order NLOS.
	We can use results from \cite{yin2022millimeter} to classify received signals accordingly.
	
	We design several EKFs governed by nonlinear observation and process models
	% 	The Extended Kalman filter (EKF) enables the estimation of a state $x\in\mathcal{R}^{l}$ while having access to a certain set of observations $z\in\mathcal{R}^{r}$. The state and observation are governed by possibly non-linear models $f$ and $h$ 
	\begin{align}
		x_{i} &= f(x_{i-1}, u_{i}) + w_{i}\quad \text{and}\quad
		z_{i} = h(x_{i}) + v_{i}
	\end{align}
	where $u$ is the control, $x$ the state and $z$ the measurement.
	Both the state and observation are subject to Gaussian white noise processes $w$ and $v$ with covariances $W$ and $V$.
	The estimate of the state $x_{i|j}$ at control iteration $i$ given observations up to time $j \leq i$ and the corresponding estimation covariance $P_{i|j}$ are updated by the Kalman equations (see App.~\ref{app:ekfgrad} for explicit computations of the derivatives of the estimation covariance) with first-order linearizations of the non-linear models $f$ and $h$.
	
	We assume a Cartesian 2D planar coordinate system and aim to identify the robot position $p$ with respect to a signal emitting transponder $p_{tx}$. Due to the directional character of the mmWave band this implies that the robot can directly infer the AoA $\alpha$ of the received signal. In the following, lower indices $a_x$ and $a_y$ represent the $x$ and $y$ component of a 2D position vector $a$. 
	
	\subsection{Line-of-sight case}
	\label{sec:LOS}
	In this case we receive the signal directly from the emitter
	without any reflection.
	We identify the transponder as the origin of the coordinate system. The robot position $p$ is chosen as the filter state $x\coloneqq p$. We assume simple integrator dynamics for the mobile robot
	\begin{equation}
		p_{i+1} =  p_i + u_i
		\label{a1}
	\end{equation}
	We then write the observation model as
	\begin{equation}
		z_{i}\hspace{-2pt} =\hspace{-2pt}
		\BIN 
		\cos(\alpha_i) &
		\sin(\alpha_i)
		\BOUT^T \quad\hspace{-10pt}\text{and}\hspace{-2pt}\hspace{-2pt}\quad \hspace{-2pt}h(x_i)\hspace{-2pt} = \hspace{-2pt}
		\BIN
		\frac{p_{x,i}}{\Vert p_i\Vert} & 
		\frac{p_{y,i}}{\Vert p_i\Vert}
		\BOUT^T
		\label{eq:obs}
	\end{equation}
	We chose a redundant representation to avoid discontinuities for example associated with the $\arctan$~\cite{Graefenstein2010}. Note however that we measure directly $\alpha_i$.
	
	\subsection{First order non-line-of-sight}
	\label{sec:NLOS}

	In the first-order NLOS case the received signal has been reflected one time.
	The reflection can be represented by
	\begin{equation}
		{(p_r - p_{tx})^T}\hat{s}/{\Vert p_r - p_{tx}\Vert} = - {(p_r - p)^T}\hat{s}/{\Vert p_r - p\Vert}
		\label{eq:sameangle}
	\end{equation}
	where $p_{tx}$ is the transponder position in some global reference system. $p_r$ is the point of reflection (POR) which lies on the line of reflection (LOR) 
	\begin{equation}
		s = \BIN 0 & c\BOUT^T + t\hat{s} \qquad \text{with }\hat{s} = \BIN 1 & m\BOUT^T
	\end{equation}
	$c$ is the offset along the $y$-axis and $m$ is the inclination of the LOR.
	The two dimensions of $p_r$ are linearly dependent
	\begin{equation}
		p_{r,y} = mp_{r,x} + c
		\label{eq:LOR}
	\end{equation}
	Together with~\eqref{eq:sameangle} we can derive a symbolic expression for $p_{r,x} = f(p,p_{tx},m,c)$. 
	The observation model becomes then (we use the same state $z_i$ as in \eqref{eq:obs})
	\begin{equation}
		h(x_i) =
		\BIN
		\frac{p_{r,x,i} - p_{x,i}}{\Vert p_{r,i} - p_i\Vert}&
		\frac{p_{r,y,i} - p_{y,i}}{\Vert p_{r,i} - p_i\Vert}
		\BOUT^T
		\label{eq:obsnlos}
	\end{equation}
	We need to distinguish two cases, depending on whether we know the line of reflection.
	\subsubsection{Unknown line of reflection}
	\label{sec:unknownlor}
	
	In case that the LOR is unknown we place the origin of the coordinate system at the transponder ($p_{tx}=0$). The filter states are
	$
	x \coloneqq 
	\BIN
	p^T&
	m&
	c
	\BOUT^T
	$. We use the same dynamics for $p$. The dynamics of $m$ and $c$ are just random noise. 
	We see that this filter will have observability issues to identify $m$ and $c$ and will need a good initial estimate of the LOR. 
	In the following we propose instead to estimate the LOR from the current state of the visual SLAM algorithm.
	% We therefore rather rely on a filter with known LOR's from SLAM as described next.
	
	\subsubsection{Known line of reflection}
	\label{sec:knownlor}

	\begin{figure}[h!]
		\centering
		\includegraphics[width=1\columnwidth]{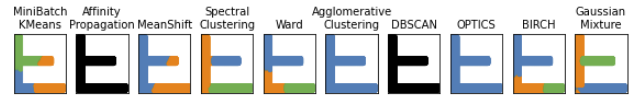}
		\vspace{-17pt}
		\caption{Clustering methods on a map with three walls. Clustering based on the Gaussian Mixture model leads to a clean separation between the different wall segments.}
		\label{fig:clustering}
	\end{figure}

	We parametrize walls by LOR's from a 2D point map of the environment (identified for example by some SLAM modality) using a clustering and regression process.
	We first identify clusters in order to separate walls from each other, for example if two walls form a corner. Figure~\ref{fig:clustering} depicts the results from a clustering process of a point cloud representing three intersecting walls using different methods. We decided to use clustering based on a Gaussian mixture model with Bayesian estimation of a Gaussian mixture~\cite{bgmm} as it led to the best separation of the walls in our experiments.
	In a second step, we use linear regression to fit the LOR's~\eqref{eq:LOR} to the identified point clusters.
	
	The filter is accordingly formulated in the global SLAM reference system. Therefore, $p_{tx}$ needs to be identified while we use the current SLAM estimate of $p$.
	The corresponding filter states are then
	\begin{equation}
		\BIN
		p_{i+1}&
		p_{tx,i+1}
		\BOUT^T
		= 
		\BIN 
		p_i + u_i&
		p_{tx,i}
		\BOUT^T
	\end{equation}
	
	\subsection{$n$-th order non-line-of-sight}
	\label{sec:NLOS_n}
	
	The first-order NLOS formulation can be extended to $n$-th order by defining $n$ LOR's $\hat{u}_k = \BIN 1 & m_k\BOUT^T$ with known reflection order $k=1,\dots,n$. For each LOR the equation  
	\begin{equation}
		\frac{(p_{r,k}-p_{r,k-1})^T}{\Vert (p_{r,k}-p_{r,k-1})\Vert}\hat{u}_k \hspace{-2pt}=\hspace{-2pt} - \frac{(p_{r,k} - p_{r,k+1})^T}{\Vert p_{r,k} - p_{r,k+1}\Vert}\hat{u}_k ,\text{ } k=1,...,n
	\end{equation}
	describes the incoming and outgoing rays at the $k$-th reflection point $p_{r,k}$. We have $p_{r,0} = p_{tx}$ (transponder position) and $p_{r,n+1}=p$ (robot position). 
	This leads to an expression for $p_{r,n,x} = f(p,p_{tx},m,c)$ with $p_{r,n,y} = m_{n}p_{r,n,x} + c_{n}$. The observation model is then the same as in~\eqref{eq:obs} with $p_{r,n}$ instead of $p_{r}$.
	
	This filter requires the knowledge of the $n$ LOR's and their correct order. 
	We wrote the filter for completeness but in practice this is an unlikely assumption. In this work, we rather reply on an exploration mode to escape such high-order reflections, see sec.~\ref{sec:alg}

	\section{Constrained path planning under uncertainty}
	\label{sec:path}
	
	We now present the trjectory optimization approach. Our goal is to compute paths
	that will move the robot towards the estimated emitter position while avoiding obstacles
	but that will additionally try to minimize the predicted uncertainty on these estimates.
	We want to find paths that will take into account estimation uncertainty to improve overall navigation.
	
	In order to receive AoA measurements that lead to EKF state estimates with minimal uncertainty we formulate the following trajectory optimization problem 
	\begin{align}
		&\mini_{x,u} \qquad \gamma_N(x_{N|N},u_N) + \sum_{i}^{i+N-1} \gamma_i(x_{i|i},u_i)\label{eq:trajopt}\\
		&x_{i|i} \coloneqq \BIN p_{i|i}^T  & \text{vec}(P_{i|i})^T\BOUT^T \leftarrow \text{EKF}(x_{i-1|i-1}, u_i)
		\label{eq:dynamics}\\
		&\phi_i(x_{i-1|i-1},u_i) \leq 0 \qquad \text{with }i = 1,...,N
	\end{align}
	This is a non-linear optimal control problem over a control horizon of length $N$ where we aim to minimize a 
	cost composed of running $\gamma_i$ and terminal $\gamma_N$ costs.
	We aim to find for each time step $i$ of the planning horizon states $x_{i|i}$ and controls $u_i$ that minimize this cost while satisfying the constraints $\phi_i$. The state $x_{i|i}$ is comprised of the robot position estimate $p_{i|i}$ and the associated vectorized estimation covariance $P_{i|i}$
	and is governed by the EKF update equations~\eqref{eq:EKFupdates} for the state and covariance as a Bayesian filtering process~\cite{berg2011}. 
	%The feedback term $Ky$ on the measurement residual $y$~\eqref{eq:EKFy} is neglected since there are no measurements along the planning horizon. 
	The transition and observation matrices are chosen according to the LOS or NLOS expressions derived in  Sec.~\ref{sec:LOS} and Sec.~\ref{sec:NLOS}. 
	
	The cost $\gamma$ incorporates the distance between estimated and desired robot position $p_{i|i}$ and $p_d$, the uncertainty $P_{i|i}$ of the position estimate $p_{i|i}$ and the control effort $u$, all subject to minimization.
	The explicit formulation for the running and terminal costs then becomes
	\begin{align}
		& \gamma_i =  u_i^T R_i u_i +  tr(P_{i|i} T_i)\\
		& \gamma_N =  (p_{N|N} - p_d)^T Q_N (p_{N|N} - p_d) +   tr(P_{N|N} T_N)\nonumber
	\end{align}
	$Q$, $R$ and $T$ are diagonal weight matrices which let us trade-off the three different objectives.
	
	We impose lower and upper bound constraints $\underline{u}$ and $\overline{u}$ on the controls.
	Different from the original formulation~\cite{berg2011}, we formulate obstacle avoidance as a constraint. This is computationally advantageous as the need for second-order derivatives in iLQR~\cite{Tassa2014} is rendered unnecessary. We have
	\begin{equation}
		\phi_{\text{obs},i}(x_{i|i}) \hspace{-2.5pt}\coloneqq  \hspace{-2.5pt} -d_{\text{euc},i}(p_{i|i})  \hspace{-2.5pt}+ \hspace{-2.5pt} r_{\text{rob}}  \hspace{-2.5pt}+ \hspace{-2.5pt} r_{\text{obs}}   \hspace{-2.5pt}+ \hspace{-2.5pt} n_{std} \max(\text{eig}( \hspace{-1pt}P_{i|i}) \hspace{-1pt})  \hspace{-2.5pt}\leq \hspace{-2.5pt} 0\label{eq:obstacle}
	\end{equation}
	$d_{\text{euc}}$ is the Euclidean distance between the robot and obstacle center which are both modeled as spheres with radii $r_{\text{rob}}$ and $r_{\text{obs}}$. $\max(\text{eig}(P_{i|i}))$ is the distance along one standard deviation and gives the constraint the character of a chance constraint $P(\phi(x,u) \leq 0) \geq 0.997$ (for $n_{std}=3$). 
	
	We use the interior point method based iLQR solver IPDDP~\cite{pavlov2021} to solve the problem~\eqref{eq:trajopt} as it enables to easily include these constraints.
	%An initial feasible trajectory from the initial position $p_{0}$ to the goal location $p_d$ is identified by A$^{*}$~\cite{Hart1968}.
	%The covariances along the planning algorithm are initialized with a weighted identity matrix.
	The solver requires the gradient of both the dynamics and the constraints $\phi$. The gradient of the Bayesian filtering can be determined analytically. The gradient of the state estimation update~\eqref{eq:EKFupdates}  is direct and depends on the chosen state transition $f(p)$. The gradient of the estimation covariance update~\eqref{eq:EKFupdates} is more involved and detailed in App.~\ref{app:ekfgrad}. 
	Finally, the analytical gradient of the obstacle avoidance constraints including the covariance term~\eqref{eq:obstacle} can be determined according to~\cite{Aa2006ComputationOE}. 
	Thanks to this formulation, we have a lower computational complexity than the original algorithm~\cite{berg2011} as summarized in Table~\ref{tab:pathplan}.
	\begin{table}[htp!]
		\centering
		\begin{tabular}{|c"c|c|} 
			\hline
			Method & $\nabla_x f(x)$~\eqref{eq:dynamics} & $\nabla^2_x \phi(x)$~\eqref{eq:obstacle} \\
			\thickhline
			Ours & $O(l^4)$ & not necessary \\
			\hline
			\cite{berg2011} & $O(l^6)$ & $O(ol^2)$ \\
			\hline
		\end{tabular}
	\vspace{-3pt}
		\caption{Computational comparison for selected operations per solver step according to~\cite{berg2011} and our method based on constrained iLQR. $o$ is the number of obstacles.}
		\label{tab:pathplan}
	\end{table}

	\section{Overall algorithm}
	\label{sec:alg}
	
	The overall algorithm consists of two steps:
	\begin{itemize}[leftmargin=*]
		\item\emph{Estimation step} updates a list of EKF's (LOS and NLOS) during the motion of the robot while observing the AoA of the received wireless signal in order to identify the approximate robot and transponder positions. New NLOS filters are added as soon as a new LOR is identified.
		\item\emph{Path planning step} re-computes an optimize path for the robot towards the goal location $p_d\coloneqq p_{tx,i|i}$ based on the current estimates of the robot and transponder positions $p_{i|i}$ and $p_{tx,i|i}$ and the obstacle map. 
	\end{itemize}
	As in our previous work \cite{yin2022millimeter}, 
	the algorithm is incorporated into the Neural-Slam~\cite{chaplot2020} system which returns a robot position estimate $p_{i|i}$ in a 2D binary point map of the environment (free space and obstacles) using vision. 
	We use the 2D map to generate LOR's. A simple heuristic is used to cover the point cloud with circles which act as a convex approximation of the obstacles for the trajectory optimization.
	
	%Since this is computationally heavy we rely on interpolation to obtain high spatial resolution AoA data as it would be available in real-world settings in order to accommodate the EKF identification process.
	
	The high-level control architecture for moving the robot to an unknown transponder position $p_{tx}$ with the current estimate $p_{tx,i|i}$ is described below. We use the machine learning based link state estimator from~\cite{yin2022millimeter} for identifying the degree of reflection of the received wireless signal.
	\subsubsection{LOS} the path planner uses the EKF formulation from Sec.~\ref{sec:LOS} for the dynamics~\eqref{eq:dynamics}. If the Mahalanobis distance~\cite{mahalanobis1936} of the filter is below a threshold, we plan a trajectory from the current robot position estimate $p_0 = p_{i|i}$ to the transponder estimate with a low weight $T$ on the covariance minimization term of~\eqref{eq:trajopt}. Otherwise we choose a high weight. 
	\subsubsection{ First-order NLOS} the path planner uses the EKF formulation from Sec.~\ref{sec:NLOS} for the dynamics~\eqref{eq:dynamics}. We check whether a virtual ray along the current received wireless signal's AoA $\alpha$ intersects with one of the identified LOR's. If this is the case its parametrization is used for the EKF. If several LOR's are intersecting we choose the closest LOR. The initial position is set to the current Neural-Slam estimate $p_0 = p_{i|i}$. The goal location is set to an offset position from the POR $p_{r}$ towards the transponder position estimate $p_{tx,i|i}$ of the EKF.
	%\begin{equation}
	%p_{d} = p_r + a\frac{p_{tx,i|i} - p_r}{\Vert p_{tx} - p_r \Vert}
	%\label{eq:fiOrNLOSpd}
	%\end{equation}
	%The scalar $a$ determines the offset from the LOR.
	This is necessary as the EKF can identify the correct distance to the LOR only if a good initial estimate exists (see Sec.~\ref{sec:evalID} below). 
	
	\subsubsection{First-order NLOS without intersecting wall, higher-order reflections, no signal} We switch to the random exploration mode presented in~\cite{chaplot2020} until the robot has approached an area with good signal reception.
	
	% The EKF formulations are formulated in 2D with an overall variable count of $6$ (two state variables and four covariance terms).
	
	\section{Evaluation}
	\label{sec:eval}
	
	We conduct  simulations using the AI Habitat robotic simulator~\cite{savva2019} with environments from the Gibson 3D indoor models~\cite{xia2018} including 3D camera data. 
	The AoA from the mmWave signals is computed offline using ray-tracing with resolution of 1~m (cf. \cite{yin2022millimeter} for more details on the wireless simulation). 
	We compute a new control trajectory every 10 control cycles.
	We choose the weights in the cost function $l$~\eqref{eq:trajopt} as $Q_N = 50I$ for the terminal state cost, $R=0.01I$ for the running control cost and $T=50I$ for the running and terminal covariance term ($T=I$ for LOS EKF with low Mahalanbois distance $<0.1$ and $T=0$ if no covariance minimization is desired). The control limit is $\bar{u}=1$m.
	
	%In this section we evaluate the methods for identification and path planning presented in the previous sections~\ref{sec:EKF} and~\ref{sec:path}.
	First, we test the LOS and NLOS EKF's for robot and transponder position identification on simple maps.
	We proceed with the evaluation of the path planner in a cluttered environment with three obstacles (sec.~\ref{sec:evalPP}). We continue with the transponder localization problem on a complex, cluttered environment. Finally, we evaluate the entire system in 200 different indoor scenarios and compare the performance with the previous system\cite{yin2022millimeter}.

	\subsection{Identification: Localization on simple maps}
	\label{sec:evalID}

	\begin{figure}
		\centering
		\vspace{4pt}
		\includegraphics[width=1\columnwidth]{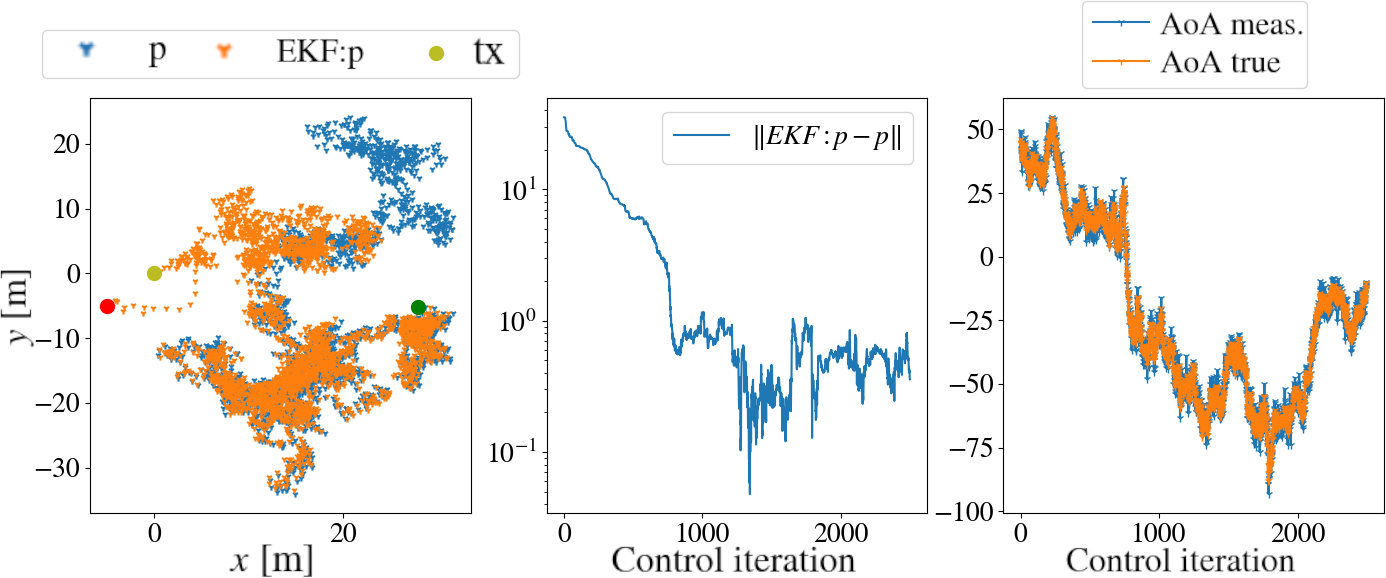}
		\vspace{-15pt}
		\caption{LOS: The filtered position $\text{EKF}\colon\hspace{-2pt} p$ (from red to green dot) converges to accuracy of approximately~1m. The robot receives a noisy measurement (meas.) of the signal's AoA (true AoA in orange).}
		\label{fig:ekf_los}%
	\end{figure}

	\begin{figure}
				\vspace{4pt}
		\centering
		\includegraphics[width=0.95\columnwidth]{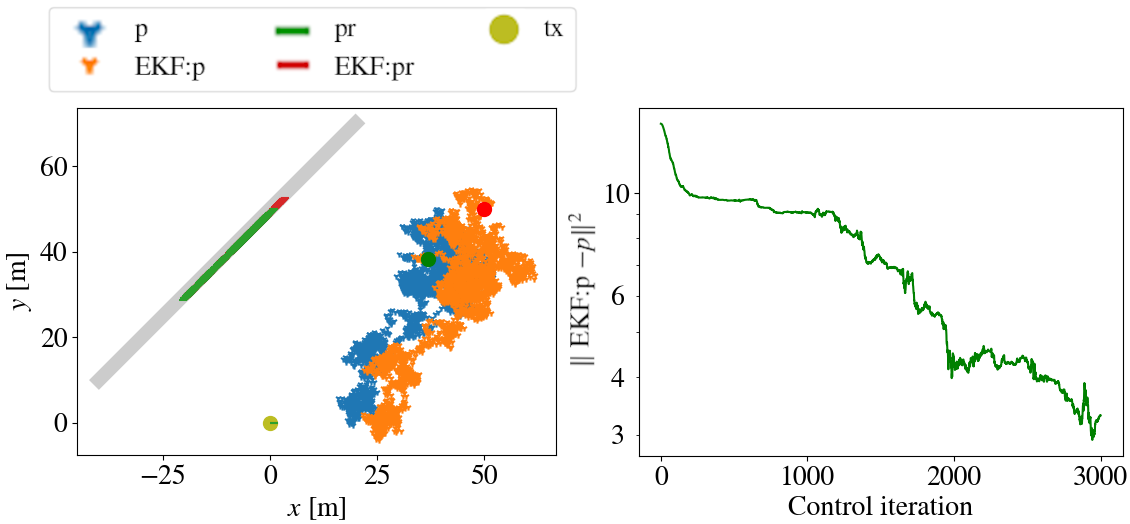}
			\vspace{-5pt}
		\caption{EKF for NLOS with known LOR. The EKF state $\text{EKF}\colon\hspace{-2pt}p$ converges to a point on the connecting line between the true POR $p_r$ and $p$.}
		\label{fig:nlos_unknown_lor}%
	\end{figure}

	The robot executes random motions in both $x$ and $y$ direction to gather AoA data with added Gaussian noise $\alpha + 0.05w$~rad with normal distribution $w \sim \mathcal{N}(0,1)$.
	The filter behavior for the LOS case is depicted in Fig.~\ref{fig:ekf_los}.  The received true and noisy AoA of the signal is given in the right upper graph. From the left and middle graphs it can be observed that the filtered robot position $\text{EKF}\colon\hspace{-2pt}p$ (orange) converges from the initial estimate with 40~m error (red) to the true one (blue) within 1~m error at the green point which corresponds to the resolution of the simulated wireless data.
	
	An EKF convergence example for the NLOS case with a known LOR is given in Fig.~\ref{fig:nlos_unknown_lor}. The position error is reduced from 10~m to 3~m.  
	The badly identified absolute distance to the POR can be explained by the high non-linearity of the observation model. Improvements may be achieved by augmenting the observations, for example with the signal strength and a corresponding signal strength decay model. Another possibility would be to use a particle filter to obtain good initial EKF estimates~\cite{Graefenstein2010}. 
	Nevertheless, as we will see later, the good directional estimates are sufficient to navigate the robot to the transponder in our experiments.

	\subsection{Evaluation of path planning}
	\label{sec:evalPP}

	\begin{figure}
				\vspace{4pt}
		\centering
		\setlength\tabcolsep{0pt} % default value: 6pt
		\begin{tabular}{cc}
			\includegraphics[width=0.5\columnwidth]{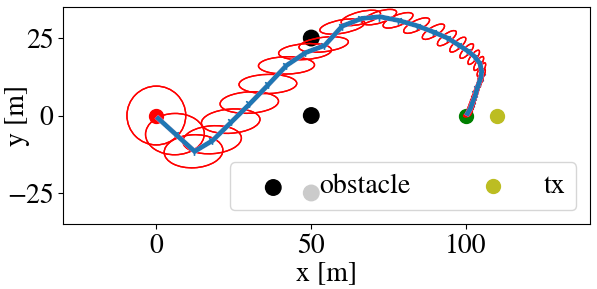} & \includegraphics[width=0.5\columnwidth]{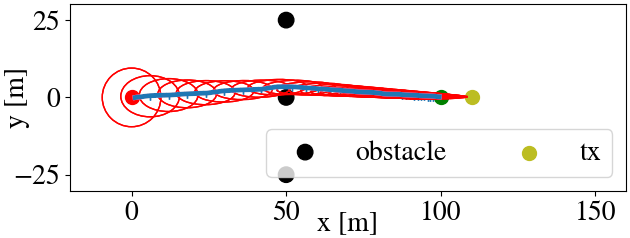}
						\vspace{-3pt}
			\\
			(a) LOS, $T=50I$ & (b) LOS, $T=0$\\
			\includegraphics[width=0.5\columnwidth]{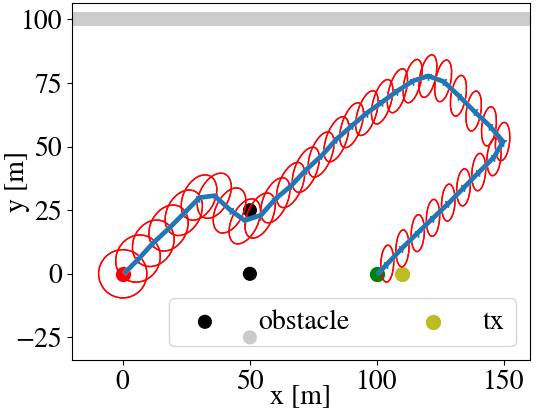} & \includegraphics[width=0.5\columnwidth]{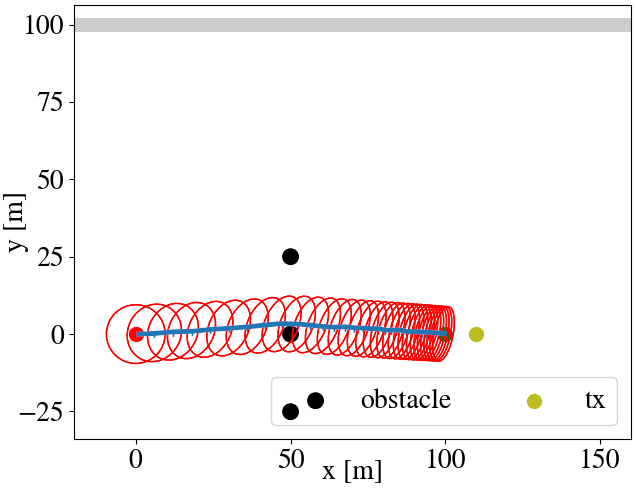}
									\vspace{-3pt}
			\\
			(c) NLOS, $T=50I$ & (d) NLOS, $T=0$
		\end{tabular}
		\vspace{-5pt}
		\caption{Computed trajectory with obstacle avoidance. The uncertainty of the state estimate is reduced significantly for the LOS case and moderately for the NLOS case as can be seen from the red confidence ellipses along the trajectories. The LOR for the NLOS case is depicted in gray.}
		\label{fig:obsctrlcov}%
	\end{figure}
	
	To illustrate how the trajectory optimization algorithm can find paths that reduce expected uncertainty, we present results in an environment with three obstacles for the LOS (sec.~\ref{sec:LOS}) and NLOS (sec.~\ref{sec:NLOS}) case (Fig.~\ref{fig:obsctrlcov}). The robot is asked to move from the origin $\BIN 0 & 0\BOUT$~m to the location $\BIN 100 & 0 \BOUT$~m while navigating three obstacles. The transponder is located at $\BIN 0 & 110 \BOUT$~m. In the NLOS case, the LOR is defined by $m=0$, $c = 100$~m. The covariance along the path of length $N=100$ is initialized as $P=0.1I$ at each collocation point such that $\Xi\coloneqq\sum_{i=1}^Ntr(P_{i|i})=20$. 
	The results on the covariance are summarized in Table~\ref{tab:obs}.
	
	For $T=50I$ the trajectory optimization results in non-trivial trajectories that minimize the uncertainty of the state estimate. In the LOS case (Fig.~\ref{fig:obsctrlcov}a), we see a clear reduction in the uncertainty of the state estimate along the trajectory. When ignoring uncertainty reduction (zero weight $T=0$ on the covariance cost function) we find a shorter path but higher uncertainty (Fig.~\ref{fig:obsctrlcov}b). For the NLOS case, the reduction is less pronounced (Fig.~\ref{fig:obsctrlcov}c) but still visible.
	We explain this by the high non-linearity of the NLOS model which makes it more difficult for the solver to find a good minimizer. In all cases, all obstacles are avoided. %The results are summarized in table~\ref{tab:obs}.
	\begin{table}[htp!]
		\centering
		\begin{tabular}{|c"c|c|c|c|c|c|} 
			\hline
			Case & $T=50I$  & $T=0I$  \\
			\thickhline LOS
			& 4.4 & 10.95\\
			\hline
			NLOS & 10.84 & 12.51\\
			\hline
		\end{tabular}
		\vspace{-3pt}
		\caption{Value $\Xi$ as the trace of the covariance along the trajectory from initial value $\Xi=20$ to after convergence of the trajectory optimization.}
		\label{tab:obs}
	\end{table}

	\subsection{Overall algorithm: Locating the transponder}
	\label{sec:evalOA}

	\begin{figure}[htp!]
		\centering
				\vspace{4pt}
		\includegraphics[width=0.6\columnwidth]{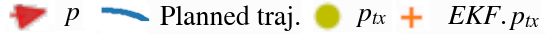}
		\\
		\setlength\tabcolsep{2pt} % default value: 6pt
		\begin{tabular}{ccc}
			\includegraphics[width=0.29\columnwidth]{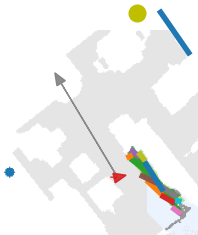} &  
			\includegraphics[width=0.29\columnwidth]{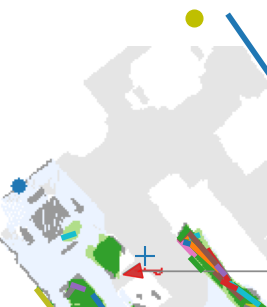} &
			\includegraphics[width=0.29\columnwidth]{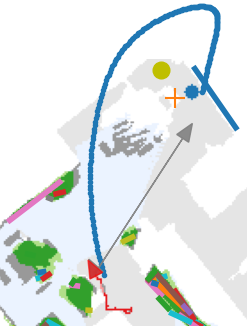}   
			\vspace{-3pt}\\
			(a) Iter. 10 & (b)  Iter. 30 & (c)  Iter. 50
			\vspace{-3pt}\\
			\includegraphics[width=0.29\columnwidth]{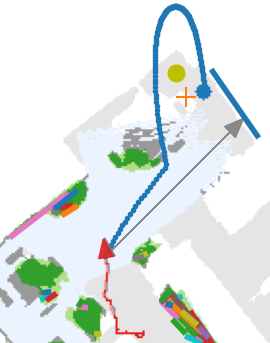}&
			\includegraphics[width=0.29\columnwidth]{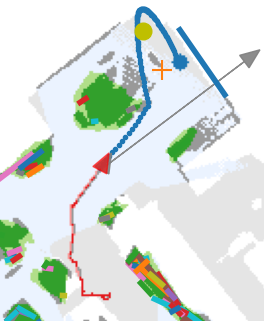} &  
			\includegraphics[width=0.29\columnwidth]{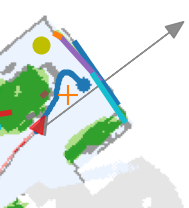} 
			\vspace{-3pt}
			\\ 
			(d)  Iter. 60 & (e)  Iter. 70  & (f)  Iter. 80 
			\vspace{-3pt}\\ 
			\includegraphics[width=0.28\columnwidth]{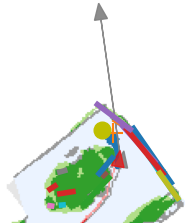} &  
			\includegraphics[width=0.255\columnwidth]{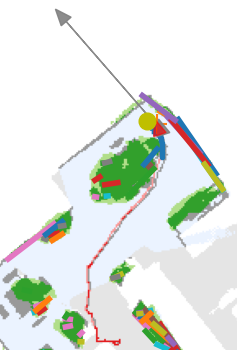}
			\vspace{-3pt}\\ 
			(g) Iter. 90 &(h) Iter. 98
		\end{tabular}
		\centering
		\vspace{-6pt}
		\caption{Path planning for transponder identification of map 'Springhill'. The point cloud identified by Neural-SLAM is shown in green and the corresponding identified LOR's are multi-colored. The direction of the received wireless signal is indicated by the gray arrow. 
			%The LOR for the first-order reflection (blue line near the transponder) is assumed to be known from the beginning.
		}
		\label{fig:pp}
	\end{figure}

	%\begin{figure}[htp!]
	%	\centering
	%	\includegraphics[width=0.9\columnwidth]{figures/fullsim6/solverdata.png}
	%	\vspace{-7.5pt}
	%	\caption{Path planning in a cluttered environment, top left to bottom right: solver time and iterations, trajectory length, number of obstacles, solver convergence and solver time normalized by the number of solver iterations and trajectory length.}
	%	\label{fig:pp_data}%
	%\end{figure}
	
	We use the complete algorithm in order to navigate the robot to a previously unknown transponder position emitting a wireless signal. To illustrate a typical behavior, we present navigation results on the Gibson map `Springhill' in Fig.~\ref{fig:pp}. Until control iteration 40 the link-state estimator correctly informs the robot that the received signal is reflected to a high order. The exploration mode following random targets is employed. 
	At control iteration 50, the received signal is a first-order reflection. The ray in direction of the AoA intersects with the blue wall (which we assume to be known by the robot from the beginning for demonstration purposes, despite being out of the camera's sensing distance). The path planner computes a trajectory with horizon length of 200 to a goal offset from the LOR. This process is repeated until the robot gets in LOS with the transponder and finally arrives at the true transponder position at control iteration 98.
	
	During the robot motion Neural-SLAM gradually constructs a point cloud (green points) of the environment. Accordingly, LOR's are fitted to it and added to the set of EKF's. The number of obstacles goes up to 300 as the robot progresses. Since the Hessian of the obstacle avoidance constraints is not necessary, this only leads to a slight increase of normalized computation times. The trajectory at control iteration 90 with 261 obstacles and a path length of 55 is computed in under 2~s with 122 solver iterations. 
	The solver converges to high precision for all trajectories and all obstacles are avoided. We also notice that the resulting motions are non-trivial as they aim to reduce state covariance.

	\subsection{Benchmark}
	\label{sec:benchmark}

	\begin{figure}[htp!]
				\vspace{4pt}
		\centering
		\includegraphics[width=1\columnwidth]{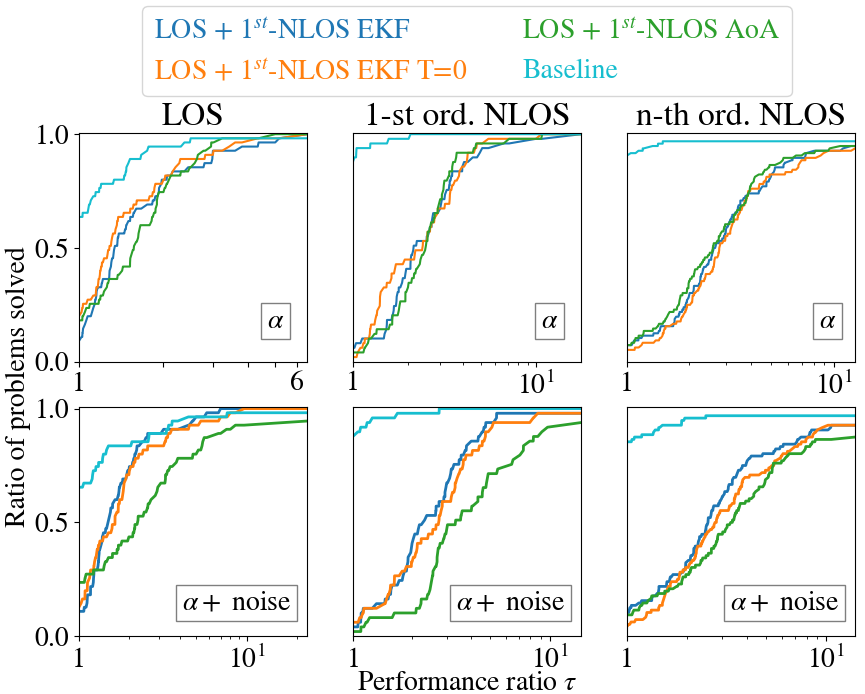}
		\vspace{-5pt}
		\caption{Performance profile for 200 transponder localization scenarios. We show the results for noise free AoA measurements (top row) and noisy AoA measurements (bottom row).}
		\label{fig:benchmark}%
		\vspace{-0.5cm}
	\end{figure}

	We now systematically evaluate the full system on 10 different maps with 10 different transponder locations each. We compare our algorithm with a baseline algorithm based on Neural-SLAM where the transponder positions are known in advance (pure navigation problem to a fixed desired goal). Furthermore, we show the results of the algorithm originally proposed in~\cite{yin2022millimeter} (LOS + 1$^{st}$-NLOS AoA) which follows a target offset in the direction of the LOS or first-order NLOS AoA. Otherwise, the algorithm depends on the same exploration mode that we employ for higher-order reflections or no signal. Our algorithm is evaluated with (LOS + 1$^{st}$-NLOS EKF) and without (LOS + 1$^{st}$-NLOS EKF T=0) covariance minimization.
	
	The results are shown in~fig.~\ref{fig:benchmark} as the ratio of problems solved over performance ratio $\tau$ (for each solver, sum all problems' path lengths to solution; each problem's path length is normalized by shortest path out of all solvers). We distinguish the cases where the robot is in LOS, first-order NLOS and higher-order NLOS in the first iteration. 
	
	In the case of noise free AoA measurements all algorithms solve all problems with initial LOS or first-order NLOS configuration with a high success rate and approximately in the same time (top row). 
	However, the pre-computed AoA arrival data from the ray-tracer can be considered `perfect' simulated data. We therefore consider the case of the AoA measurements being subject to additional white noise ($w \sim \mathcal{N}(0,1)$, $\alpha + 0.35w$~rad). The results are given in the bottom row of fig.~\ref{fig:benchmark} and summarized in table~\ref{tab:benchmark}.
	\begin{table}[htp!]
		\centering
		\begin{tabular}{|c"c|c|c|c|c|c|} 
			\hline
			Method & Success rate & Duration \\
			\thickhline
			LOS+1$^{st}$-NLOS EKF (Ours) & 94.5\% & 76\%\\
			\hline
			LOS+1$^{st}$-NLOS EKF T=0 (Ours) & 94.5\% & 82\%\\
			\hline
			LOS+1$^{st}$-NLOS AoA~\cite{yin2022millimeter} & 89.5\% & 100\%\\
			\hline
		\end{tabular}
		\vspace{-3pt}
		\caption{Summary of success rate and duration (overall path length to solution) of our method with respect to~\cite{yin2022millimeter} for noisy AoA measurements.}
		\label{tab:benchmark}
	\end{table}
	The EKF based algorithms solve the problems both with higher success rate ($94.5\%$ of problems solved for EKF based algorithms, $89.5\%$ for AoA based algorithm) and faster arrival times / shorter overall path lengths from initial point to transponder (problems solved in $76\%$ and $82\%$ ($T=0$) of the time required for the AoA based algorithm). This can be explained by the robot efficiently following a filtered AoA direction. A slight advantage in arrival times can be identified for the EKF based algorithm with covariance minimization ($92.7\%$ of the time required for the EKF based algorithm with $T=0$) as the more expressive motions enable a faster transponder localization under noisy observations.

	\section{Conclusion}

	In this paper we proposed a method for navigating robots through cluttered environments to quickly find a wireless (mmWave) emitter. Our method consists of an estimator for the emitter location and a trajectory optimization algorithm that aims to minimize estimation uncertainty while navigating the robot. Experiments demonstrate the advantages of the approach under noisy measurements. The estimator improves navigation success rate while the trajectory planner based on belief-space dynamics improves tasks completion speed.
	Additionally, by formulating a constrained trajectory optimization problem with analytic gradients our algorithm is computationally efficient and can be used in an online setting.
	Future work will include experiments on real robots.
	
	%in real world settings and make them robust with regard to challenges like higher level of disturbances for example from reflections of more detailed environments that can not be simulated due to computational limitations. 
	%On the other hand we can omit supporting tools like measurement interpolations that were necessary to overcome low spatial resolutions of simulated ray-tracing. 
	%At the same time we need to implement methods for higher accuracy transponder estimation for more accurate navigation which was not necessary in the low resolution simulation environment.

	\begin{appendices}

		\section{Partial derivatives of estimation covariance update of the EKF}
		\label{app:ekfgrad}
		\vspace{-2.5pt}
		
		In the following we determine the partial derivatives of the estimation covariance $P_{i|i}\in\mathcal{R}^{l,l}$ updates of the EKF with respect to the previous estimated state $x_{i-1|i-1}\in\mathcal{R}^{l}$ and the previous estimation covariance $P_{i-1|i-1}$ 
		\begin{align}
			{\partial P_{i|i}}/{\partial x_{i-1|i-1}} \qquad \text{ and }\qquad{\partial P_{i|i}}/{\partial P_{i-1|i-1}}
			\label{eq:dPdp}
		\end{align}
		These are necessary to solve~\eqref{eq:trajopt} by our constrained iLQR solver. The index $i$ is the time step, for example along the control horizon.
		The EKF update equations are given by
		\begin{align}
			x_{i|i} \hspace{-1pt}=\hspace{-1pt} f(x_{i-1|i-1}, u_i) \hspace{-1pt}+\hspace{-1pt} K_i y_i \hspace{-1pt}\quad \hspace{-2pt}\text{and}\hspace{-2pt}\quad\hspace{-1pt}
			P_{i|i} \hspace{-1pt}=\hspace{-1pt} (I \hspace{-1pt}-\hspace{-1pt} K_iH_i)P_{i|i-1}
			\label{eq:EKFupdates}
		\end{align}
		with
		\begin{align}
			y_i = z_i - h(x_{i|i-1})\label{eq:EKFy}
			&\quad \text{and} \quad
			P_{i|i-1} = F_i P_{i-1|i-1}F_i^T + Q_i\nonumber\\
			S_i = H_i\zeta_i + R_i&\quad \text{and} \quad
			K_i = \zeta_iS_i^{-1}\nonumber\\
			F = \nabla_x f\in\mathcal{R}^{l,l} &\quad \text{and} \quad
			H = \nabla_x h\in\mathcal{R}^{r,l}
		\end{align}
		%Following relationships hold~\cite{Petersen2012, Brookes2020}:
		%\begin{align}
		%&\frac{\partial AXB}{\partial X} = B^T \otimes A\quad \text{and} \quad
		%\frac{\partial AX^TB}{\partial X} = \left(A \otimes B^T\right)\Pi\\
		%&\frac{\partial A X^{-1} B}{\partial X} = - B^TX^{-T}  \otimes A X^{-1}\\
		%&\frac{\partial A X^{-T} B}{\partial X} = - \left(AX^{-T} \otimes B^T X^{-1}\right)\Pi
		%\end{align} 
		%$X$ is a full-rank square matrix. $A$ and $B$ are square matrices.
		%$\Pi$ is the commutation matrix for Kronecker products.
		%\begin{equation}
		%\Pi = \BIN 1 & 0 & 0 & 0 \\
		%0 & 0 & 1 & 0\\
		%0 & 1 & 0 & 0\\
		%0 & 0 & 0& 1\BOUT
		%\end{equation}
		%The product rule applies. The chain rule for matrix functions is
		%\begin{equation}
		%\frac{\partial G(U(x))}{\partial x} = \left(\frac{\partial G}{\partial U}\right)^T \frac{\partial U}{\partial x}
		%\end{equation}
		%The matrix $G$ is a function of the matrix $U(x)$ which again is dependent of the vector $x$.
		We define $\zeta_i \coloneqq P_{i|i-1}H_i^T$ with symmetry of $P_{i|i-1}$.
		
		We use the relationships for matrix derivatives given  in~\cite{Petersen2012, Brookes2020}. $\Pi$ is the commutation matrix for Kronecker products. The product rule and chain rule for matrix functions applies.
		For the partial derivative w.r.t. the state estimate we get
		\begin{align}
			&{\partial P_{i|i}}/{\partial x_{i-1,i-1}} = - \left(P_{i|i-1} \otimes \left(\zeta_i S_i^{-1}\right)^T \right. \label{eq:ekfpd1}\\
			&+ \left(P_{i|i-1}^T \otimes S^{-1}\zeta_i^T \right)\Pi- \zeta_i S_i^{-1} \zeta_i^T \otimes \left(\zeta_i S_i^{-1}\right)^T\nonumber\\
			&\left.-\left(\left(\zeta_i S_i^{-1} \zeta_i^T\right)^T \otimes S_i^{-1}\zeta_i^T\right)\Pi \right)^T {\partial H_i}/{\partial x_{i-1|i-1}}\nonumber
		\end{align}
		%The partial derivative $\partial H_i / \partial p_{i-1|i-1}$ is dependent on the chosen observation model $h(x)$ and can be determined analytically or by finite differences, dependent on the complexity of the model.
		For the partial derivative w.r.t to the previous estimation covariance we have 
		\begin{align}
			&{\partial P_{i|i}}/{\partial P_{i-1|i-1}} = 
			F_i^T \otimes F_i^T- \left(F_i^T \otimes F_i^T (\zeta_i S_i^{-1}  H_i)^T\right.-\nonumber\\
			& (H_i F_i)^T S_i^{-1} \zeta_i^T \hspace{-2pt} \otimes \hspace{-2pt}(\zeta_i S_i^{-1}  (H_i  F_i))^T\left.\hspace{-2pt}+ (\zeta_i  S_i^{-T} H_i F_i)^T \hspace{-2pt}\otimes\hspace{-2pt} F_i^T\right)\label{eq:ekfpd2}
		\end{align}
		
		The partial derivatives are  computed in approximately $O(l^6)$ operations using central differences~\cite{berg2011}. The analytic computation is of order $O(8l^4 + 13lr^2 + 14l^2r + 2r^3 + 3l^3)$. The leading factor for finite differences is therefore $O(l^6)$ whereas for the analytic expression it is only $O(l^4)$ (assuming $l>r$).
	\end{appendices}

		\balance
	\bibliographystyle{IEEEtran}
	\bibliography{bib}

\end{document}